\documentclass[conference]{IEEEtran}
\IEEEoverridecommandlockouts
\def\BibTeX{{\rm B\kern-.05em{\sc i\kern-.025em b}\kern-.08em
    T\kern-.1667em\lower.7ex\hbox{E}\kern-.125emX}}
\usepackage{url,amsmath,amsfonts} 
\usepackage{algorithmic}
\usepackage{graphicx,multirow}
\usepackage{float} 
\usepackage{xcolor}
\graphicspath{{figs/}}
\usepackage{footmisc}

\begin{document}

\title{Domain Transformer: Predicting Samples of Unseen, Future Domains}

 \author{\IEEEauthorblockN{Johannes Schneider}
 \IEEEauthorblockA{\textit{Institute of Information Systems} \\
 \textit{University of Liechtenstein}\\
 Vaduz, Liechtenstein \\
 johannes.schneider@uni.li}} 

%

%
\maketitle              
\begin{abstract} 
The data distribution commonly evolves over time leading to problems such as concept drift that often decrease classifier performance. Current techniques are not adequate for this problem because they either require detailed knowledge of the transformation or are not suited for anticipating unseen domains but can only adapt to domains, where data samples are available. We seek to predict unseen data (and their labels) allowing us to tackle challenges s a non-constant data distribution in a \emph{proactive} manner rather than detecting and reacting to already existing changes that might already have led to errors.
To this end, we learn a domain transformer in an unsupervised manner that allows generating data of unseen domains.
Our approach first matches independently learned latent representations of two given domains obtained from an auto-encoder using a Cycle-GAN. In turn, a transformation of the original samples can be learned that can be applied iteratively to extrapolate to unseen domains. Our evaluation of CNNs on image data confirms the usefulness of the approach. It also achieves very good results on the well-known problem of unsupervised domain adaption, where only labels but no samples have to be predicted. Code is available at \url{https://github.com/JohnTailor/DoTra}.


\end{abstract}


\section{Introduction}
\begin{figure}
 \centering{  \includegraphics[width=\linewidth]{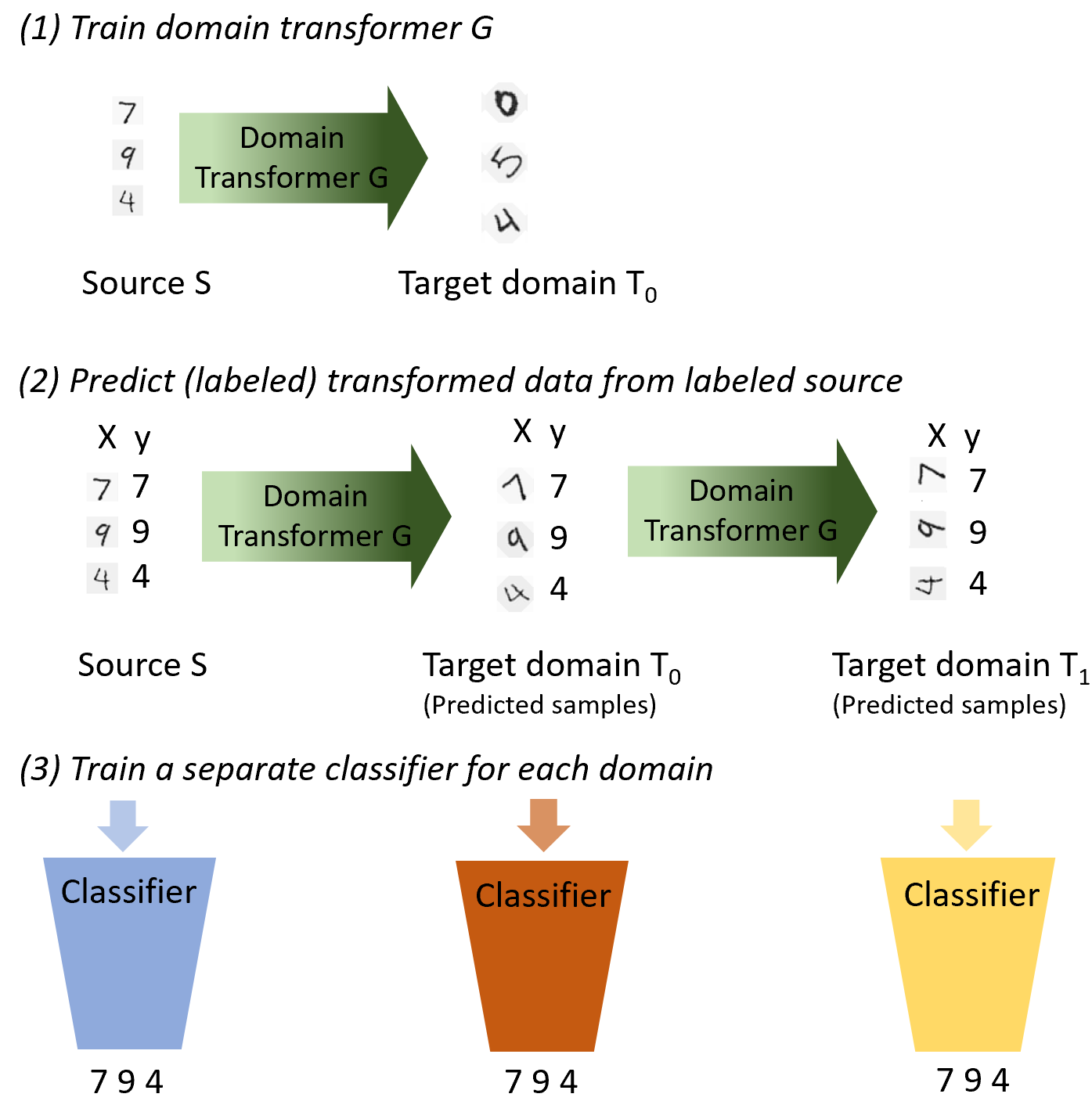}}
 \caption{Domain transformation to predict samples of ``future'' domains (target domain $T_1$ in image) given a source and target domain} \label{fig:idea}
\end{figure}

For supervised learning, training and test data should originate from the same distribution. If this condition is violated, the performance of a machine learning model typically deteriorates to the point that it is of no value. In practice, data distribution is often non-static being subject to concept drift. Current approaches aim to detect concept drift, addressing it after it occurred\cite{lu18}. Once a change in data distribution has been identified, approaches of unsupervised domain adaption can be used to adjust to the new data distribution. Unsupervised domain adaption seeks to elevate the problem of data labeling due to varying data or label distribution given labeled data of a source domain and unlabeled data of a target domain. Thus, \emph{after} a change has occurred domain adaption seeks to react to these changes, e.g., in a semi-supervised manner by trying to correctly label samples in the target domain. Labeled data can then be used to retrain the classifier. Such a reactive approach is likely to cause misclassifications since models are only adjusted after the data distribution has changed. In our work, we propose a proactive, predictive approach. That is, we assume that the data distribution evolves in a predictable, but unknown manner. For example, physical aging can alter material properties governed by some hidden transformational laws making objects look and behave differently over time. In turn, a computer vision model trained on the original model might fail to recognize objects changed due to the aging process. \\
The ultimate goal of domain prediction is to  classify ``future'' data differing from training data through an unknown transformation, i.e., the data (both samples and labels) cannot even be captured when a classifier is trained. We learn a transformer between domains given a source and (unlabeled) target domain. It can be applied iteratively to extrapolate to unseen domains. Specifically, it can transform samples from the labeled source data into (labeled) data in any target domain. These transformed samples share the same label as those in the source domain and can be used for training a classifier in unseen domains. This is illustrated in Figure \ref{fig:idea}) where neither labeled nor unlabeled samples from target domain 1 are yet available, but they can be generated using repeated transformer application.  

To learn the transformation, our idea is to encode data from the given domains to facilitate the mapping between the two domains (see Figure \ref{fig:arch}). Once we have a mapping between latent spaces obtained with a Cycle-GAN, we can construct paired samples that can be further used to train a transformer.  Our approach of transforming between encodings of different domains overall outperforms a number of other methods. However, to our surprise, a well-adjusted ``standard'' Cycle-GAN often outperforms numerous methods proposed in the literature. Thus, we investigate failure modes of Cycle-GANs, concluding that a narrow (though sufficient) receptive field and overlapping domains are inhibitors for Cycle-GANs, but less so for our proposed architecture. We also find that our architecture is easier to train, i.e., less prone to mode collapse.
Our architecture can also yield good results in the classical setting of unsupervised domain adaption, where the goal is to infer labels of unlabelled samples of a target domain.

\section{Problem} 
Let $D_S=\{(X,Y)\} \subseteq S$ denote labeled samples from the source domain $S$. We consider a sequence of target domains $T_i=\{X\}$ consisting of unlabeled samples. Let $D_{T_0}=\{X\} \subseteq T_0 $ be the given samples of the first target domain $T_0$. We assume that domains are transformable via an unknown domain transformation $G^u$, i.e. if $X \in S$ then $G^u(X) \in T_0$ and if $X' \in T_i$ then $G^u(X') \in T_{i+1}$. We call the  \emph{k-domain transformation identification (DTI)} problem, the problem of estimating $G^u$ given $D_S$ and $D_{T_i}$ with $i \in [0,k-1]$. 

In this work, we focus on $k=1$, i.e., we are given $D_S$ and $D_{T_0}$. Having access to more domains (larger $k$) might allow learning more complex time-depending evolution of data. But the case $k=1$ is most interesting, since acquiring data of many domains can be tedious, and it is likely to require more time, delaying the application of an automatic approach. To ensure error-free operation, it suffices to anticipate the next upcoming domain. If a classifier is exposed to domain $i$ at time $i$ then predicting the target domain $T_{i+1}$ given $T_i$ and $T_{i-1}$ suffices to guarantee correct operation.

\begin{figure}[!h]
 \centering{  \includegraphics[width=0.95\linewidth]{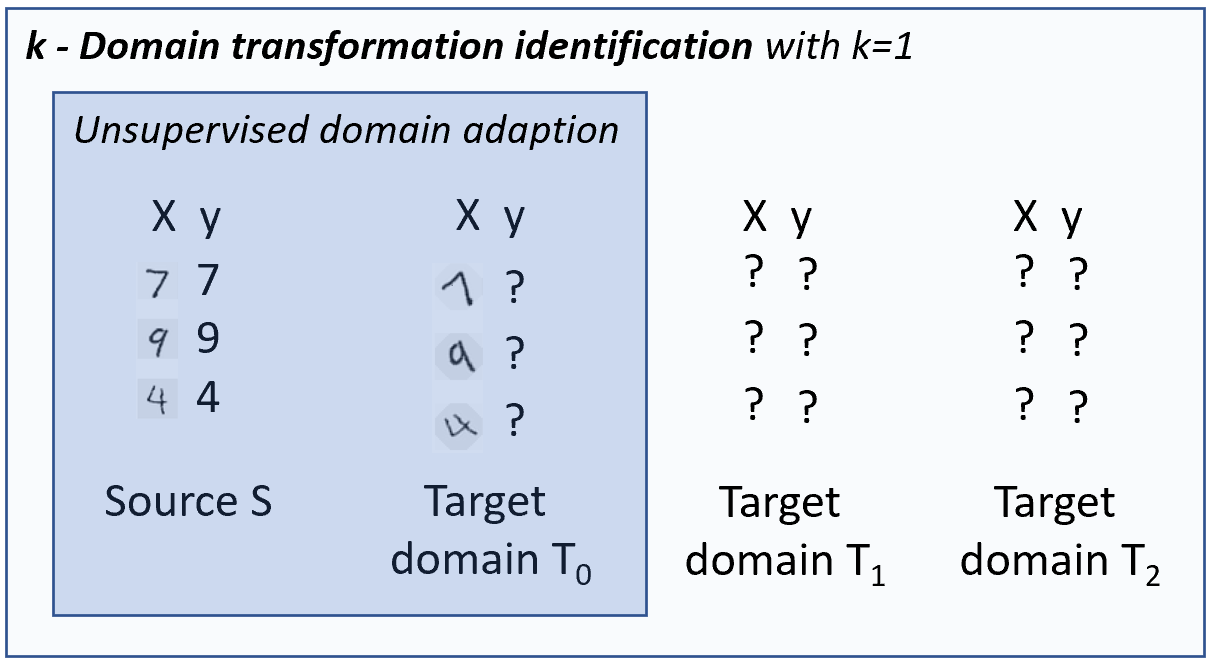}}
 \caption{Comparison of given data and data to predict (indicated by questions marks `?') for ``classical'' unsupervised domain adaption and our k-domain transformation identification problem (DTI). For DTI also samples must be predicted. For $k=1$ we only get samples from target domain $T_0$ and must predict domains $T_1,T_2,\ldots$} \label{fig:prob}
\end{figure}

A solution to the DTI problem is also a solution to the unsupervised domain adaption problem, where one seeks to predict labels of samples of $D_{T_0}$ given the labeled data $D_S$ from the source domain and the samples $D_{T_0}$. If a transformation $G$ has been identified, one can simply transform the samples $D_S$ to domain $T_0$ and train a classifier in the new domain. This classifier can then label samples $D_{T_0}$ as illustrated in Figure \ref{fig:idea}. A comparison  between unsupervised domain adaption and our $k$-domain transformation identification problem stating also given and non-given data is shown in Figure \ref{fig:prob}.


\section{Approach}
We leverage two high-level ideas. The first is to learn a mapping between latent spaces of the source domain and the given target domain. The second is to create paired samples and learn a transformer on paired samples, which allows getting a high-quality transformer more easily than on unpaired samples.\\
\begin{figure*}[!h]
 \centering{  \includegraphics[width=0.7\linewidth]{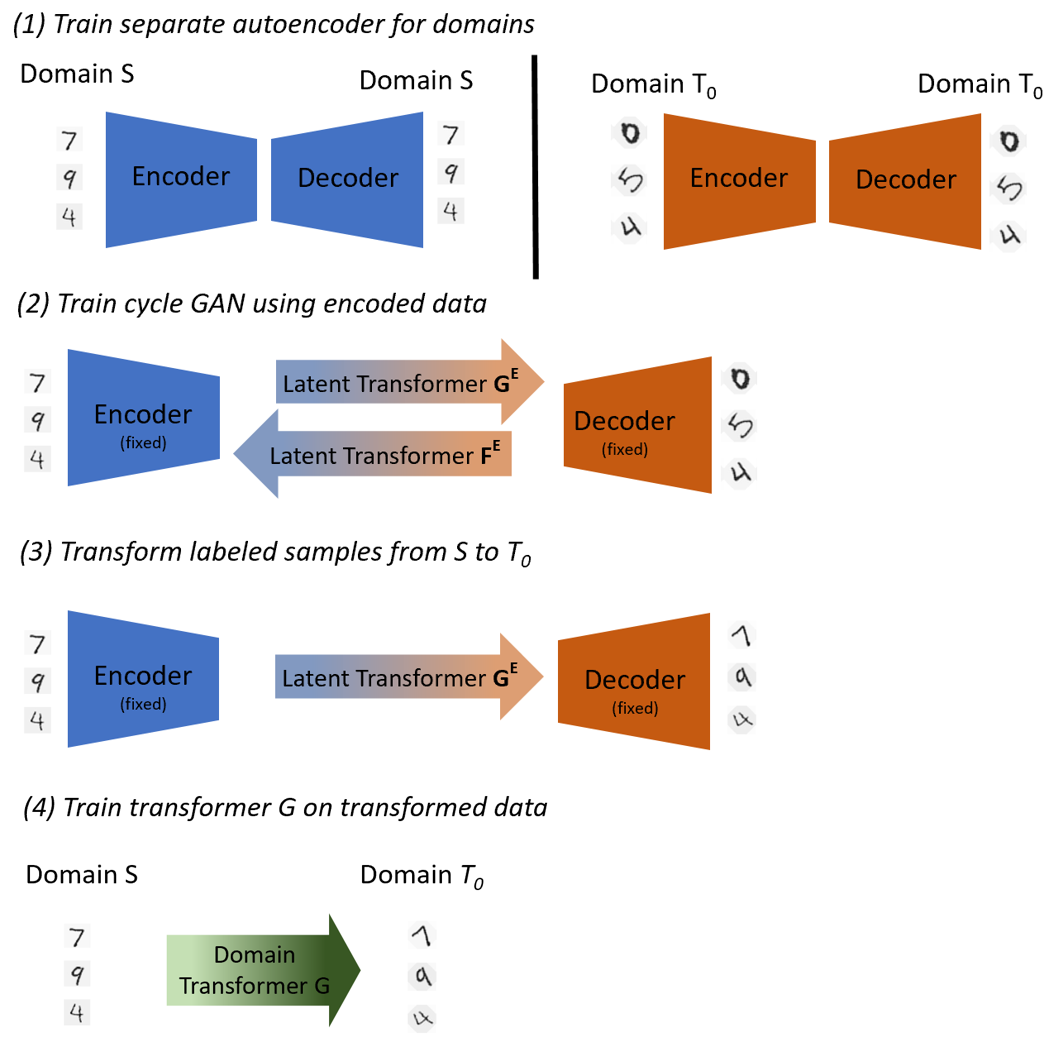}}
 \caption{Architecture and training of the Domain Transformer (DoTra)} \label{fig:arch}
\end{figure*}

The literature already provides a solution to the $k$-domain transformation identification (DTI) problem for $k=1$, i.e., Cycle-GANs\cite{zhu17}. Cycle-GANs transfer data between two domains by translating samples from one domain to the other using a generator, serving as a domain transformer. While the learned transformer can be used to extrapolate to unseen domains, it sometimes leads to poor outcomes, as we shall show and discuss in our evaluation. To improve performance, we aim at a simpler problem than translating directly from the source to the target domain: We perform dimensionality reduction in both domains (See Step (1) in Figure \ref{fig:arch}). We only map dimensions of latent spaces of different domains using the Cycle-GAN. To obtain the latent spaces, we train autoencoders (AE). During the training of the Cycle-GAN the AEs, i.e., the encoders and decoders, remain fixed (See Step (2) in Figure \ref{fig:arch}). The Cycle-GAN yields a generator, i.e., a transformer, that allows converting samples from one latent domain to another to do the inverse transformation. However, neither of these transformers operating on latent codes helps extrapolate to unseen domains, since we lack a decoder to turn the transformed (latent) codes into samples. Thus, our approach is to obtain paired samples of the source and (first) target domain and to learn a domain transformer between paired samples. This problem is often significantly easier than transforming between domains using unpaired samples. To obtain the paired samples (See Step (3) in Figure \ref{fig:arch}), we use the labeled samples from the source domain, encode them, use the transformer of the Cycle-GAN to translate them to latent codes of the first target domain, and decode the latent code. Thus, we have not only paired samples, but these paired samples also have labels, i.e., we are given the label of the source domain, and the paired sample in the target domain is simply assigned the same label. We learn a transformer on the paired samples (See Step (4) in Figure \ref{fig:arch}). The transformer trained on paired samples can be used iteratively to obtain samples from unseen domains (See Step (3) in Figure \ref{fig:idea}).\\
\noindent\textbf{Formalization:}
Next, we provide a more formal treatment for the steps in Figure \ref{fig:arch}. For Step (1) we train autoencoders. An AE $=(EN_D,DE_D)$ for domain $D$ is given by an encoder $EN_D$ and decoder $DE_D$. The L2-loss for a sample $X$ is $$L_{AE}(X,EN_D,DE_D):=||DE_D(EN_D(X))-X||^2$$  We train two AEs, one for the source domain $(EN_{S},DE_{S})$ and one for the first (and only given) target domain $(EN_{{T_0}},DE_{{T_0}})$. The AEs remain fixed for all later steps. For Step (2) we apply a Cycle-GAN with a discriminator $DI$ on the encoded samples from both domains to learn mappings $G^E$ and $F^E$ between encoded domains $EN_{S}(D_S)$ and $EN_{{T_0}}(D_{T_0})$. 
The Cycle-GAN the loss $L$ is as in \cite{zhu17} but computed on encoded data:
\begin{align}
&L(G^E,F^E,EN_{S}(D_S),EN_{{T_0}}(D_{T_0}) := \nonumber \\  
&\quad L_{GAN}(G^E,DI_{D_{T_0}},EN(D_{S}),EN(D_{T_0})) \nonumber \\  &\quad+L_{GAN}(F^E,DI_{D_{S}},EN(D_{T_0}),EN(D_{S})) \nonumber\\
&\quad+ \lambda\cdot  L_{cyc}(G^E,F^E)\nonumber \\
&\text{ with } L_{GAN}(G,DI,X,Y):= \mathbf{E}_{Y\sim p_{data}(Y)}[\log DI(Y)]  \nonumber\\
&\text{\phantom{abc}} + \mathbf{E}_{X\sim p_{data}(X)}[\log (1-DI(G(X)))] \nonumber \\
&\text{ with } L_{cyc}(G,F):= \mathbf{E}_{X\sim p_{data}(X)}[||F(G(X))-X||_1] \nonumber\\
&\text{\phantom{abc}}+\mathbf{E}_{Y\sim p_{data}(Y)}[||G(F(Y))-Y||_1] \nonumber
\end{align}

In Step (4) in Figure \ref{fig:arch} we train a transformer $G$ to map from source to the given target domain. It uses a source sample and the transformation of the source sample to the target domain, using the encoder $EN_{{T_0}}$, the latent transformer $G^E$ and the learned decoder on the target domain $DE_{{T_0}}$, i.e. the transformation of sample $X \in D_S$ to domain $T_0$ is $DE_{{T_0}}(G^E(EN_{S}(X)))$. 
That is for the domain transformer (DoTra), the loss is $$L(G):=(G(X)-DE_{{T_0}}(G^E(EN_{S}(X))))^2$$

We apply the transformer $G$ iteratively to predict samples from unseen domains. (We cannot use the latent transformer $G^E$, since it is tied to the specific domain it was trained on, i.e., we lack a decoder if we apply it to get encoded samples for $D_{T_i}$ with $i>0$.)




\section{Evaluation}
We perform an ablation study to understand the benefits of DoTra compared to using a ``standard'', direct Cycle-GAN (StaCyc) to learn the transformer between domains directly. We also compare against multiple other methods that were designed with domain extrapolation in mind.\\ 
\noindent\textbf{Architecture:} 
We employed a simple classifier to assess predictive performance, e.g. a VGG-6 network consisting of 5 repetitions of CRBM (conv-relu-batchnorm-maxpool) layers followed by a dropout and linear layer. For the AE, the encoder comprises 5 conv-batchnorm-leakyrelu (CBL) layers, where the conv-layer used stride 2 and kernel width 4 and the leak was 0.05. The final layer was a linear layer. The decoder is analogous to the encoder, but it replaces the conv layers with deconv layers for upsampling, i.e., the layer sequence is DBL instead of CBL. The Cycle-GAN's generator and discriminator, used to transform between latent spaces, both consist of two linear layers followed by batchnorm and leakyrelus with leak 0.05. 
For the domain transformer (DoTra) we used the architecture, which  is based on a common repository for Cycle-GANs \footnote{\url{https://github.com/yunjey/mnist-svhn-transfer/} \label{foot1}}. 
It is shown in Table \ref{tab:arch}. 

 \begin{table}[ht] 	
 	\caption{DoTra Architecture. C is a conv layer and D a deconv layer; A BatchNorm and LeakyReLU layer(0.05) followed each layer;}  \label{tab:arch} 
 	\begin{center}
 		\scriptsize
 		\setlength\tabcolsep{2.5pt}
	\centering
 		\begin{tabular}{|l| l |   }\hline
			Type/Stride& Filter Shape \\  \hline
			  C/s2     & $4\tiny{\times} 4 \tiny{\times} 1 \tiny{\times} 64$ \\ \hline
 			  C/s2     &$4\tiny{\times} 4 \tiny{\times} 64 \tiny{\times} 128$ \\ \hline
 			  C/s1     &$3\tiny{\times} 3 \tiny{\times} 128 \tiny{\times} 128$ \\ \hline
 			  C/s1     &$3\tiny{\times} 3 \tiny{\times} 128 \tiny{\times} 128$ \\ \hline
 			  D/s2     & $4\tiny{\times} 4 \tiny{\times} 128 \tiny{\times} 64$ \\ \hline
 			  D/s2     &$4\tiny{\times} 4 \tiny{\times} 64 \tiny{\times} 1$ \\ \hline
 			\end{tabular}
 	\end{center}
 	
 \end{table}

\begin{figure*}
 \centering{  \includegraphics[width=0.7\linewidth]{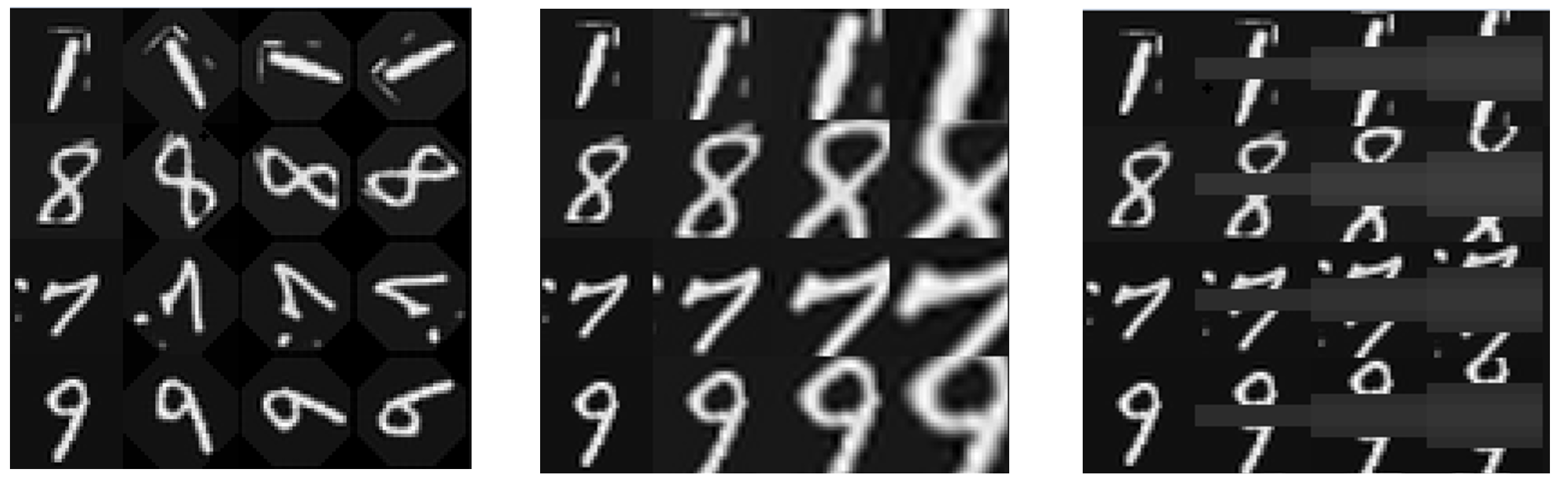}}
 \caption{Domain transformations: rotation, zoom, and split. Columns indicate domains, i.e. source $S$, target domains $T_0$, $T_1$, and $T_2$.} \label{fig:Trans}
\end{figure*}

In particular, we use the same architecture for the generator of a ``standard'' Cycle-GAN (StaCyc) in our ablation study. Code is available at \url{https://github.com/JohnTailor/DoTra}.\\
\noindent\textbf{Data and Transformations:} Aligned with prior work, we evaluate on the \emph{MNIST} dataset and, additionally, the Fashion(-MNIST) dataset. We scaled both from 28x28 to 32x32. While prior work focuses on rotation, we also use zoom and splitting in the middle (\emph{SplitMid}). We assume that the unknown transformation $G^u$ is either a 45 degree rotation,  zooming in at the center by 1.33 (and cropping to remain the original size), or a split in the middle by adding 6 pixels (and cropping). Examples are in Figure \ref{fig:Trans}. 

\begin{table}[!htb]
    \setlength\tabcolsep{3.5pt}
	\centering
	\caption{Results MNIST: Overall, DoTra performs best on the target domains $T_0$ and $T_1$. Note, DoTra predicts samples for target domains $T_1$ and $T_2$, while other methods require samples to be given.}\label{tab:results} 
	\scriptsize
	\begin{tabular}{|r| r|r|r| r| c|	} \hline 
Method/& Source           & Target& Target & Target & \footnotesize{\textbf{Given Target}}   \\
\emph{Operation}&  Domain &Dom. 0& Dom. 1&Dom. 2 & \footnotesize{\textbf{Domains}}\\ \hline \hline 
\multicolumn{1}{|l|}{\emph{Zoom}}&\multicolumn{1}{|l|}{\emph{ 0 }}&\multicolumn{1}{|l|}{\emph{ 1.33 }}&\multicolumn{1}{|l|}{\emph{ 1.77 }}&\multicolumn{1}{|l|}{\emph{ 2.35}} & \\ \hline
\textbf{DoTra}\tiny{(Our)} &99.5&99.3&96.6&64.8&\multirow{2}{*}{\small{\textbf{only 0}}}\\ \cline{1-5}
StaCyc& 99.5&98.7&87.7&58.1&\\ \hline
Source&99.6&95.9&50.2&25.7&-\\ \hline
ADDA&98.8&97.4&89.0&64.8&\multirow{5}{*}{\small{\textbf{0, 1 and 2}}}\\ \cline{1-5}
CUA&97.1&95.5&94.8&85.9&\\ \cline{1-5}
DANN&98.6&91.5&73.5&56.8&\\ \cline{1-5}
PCIDA&98.3&84.1&62.6&46.7&\\ \cline{1-5}
SO&99.2&80.0&25.2&9.7&\\ \hline \hline

\hline \multicolumn{1}{|l|}{\emph{SplitMid}} &\multicolumn{1}{|l|}{\emph{ 0 }}&\multicolumn{1}{|l|}{\emph{ 6 }}&\multicolumn{1}{|l|}{\emph{ 12 }}&\multicolumn{1}{|l|}{\emph{ 18}} & \\ \hline
   \textbf{DoTra}\tiny{(Our)} &99.5&99.1&94.8&71.7&\multirow{2}{*}{\small{\textbf{only 0}}}\\ \cline{1-5}
   StaCyc& 99.5&98.9&97.6&85.1&\\ \hline
   Source&99.6&77.3&41.8&25.8&-\\ \hline
ADDA&98.7&92.2&52.4&20.4&\multirow{5}{*}{\small{\textbf{0, 1 and 2}}}\\ \cline{1-5}
CUA&96.3&90.9&88.7&83.2&\\ \cline{1-5}
DANN&98.5&94.6&92.2&85.6&\\ \cline{1-5}
PCIDA&98.3&42.1&13.0&14.4&\\ \cline{1-5}
SO&99.2&52.3&14.9&8.9&\\ \hline \hline

\hline \multicolumn{1}{|l|}{\emph{Rotation}} &\multicolumn{1}{|l|}{\emph{ 0 }}&\multicolumn{1}{|l|}{\emph{ 45 }}&\multicolumn{1}{|l|}{\emph{ 90 }}&\multicolumn{1}{|l|}{\emph{ 135 }} &\\ \hline
\textbf{DoTra}\tiny{(Our)} &99.5&99.0&78.4&42.4&\multirow{2}{*}{\small{\textbf{only 0}}}\\ \cline{1-5}
StaCyc& 99.5&47.4&21.9&13.6&\\ \hline
Source&99.6&69.5&17.5&32.6&-\\ \hline
ADDA&98.6&71.2&16.4&26.0&\multirow{5}{*}{\small{\textbf{0, 1 and 2}}}\\ \cline{1-5}
CUA&94.6&59.5&41.6&32.4&\\ \cline{1-5}
DANN&97.8&74.3&22.6&18.7&\\ \cline{1-5}
PCIDA&98.8&83.5&99.1&97.1&\\ \cline{1-5}
SO&99.4&54.2&13.8&21.5&\\ \hline

\end{tabular}
	
	
\end{table}

\noindent\textbf{Setup:} We trained on PyTorch 1.9.0, Python 3.8 on an Ubuntu machine with an NVIDIA RTX 2080 TI GPU. For comparison to other methods, we used a classifier trained on source data $D_S$ as a baseline. To assess the impact of our modification of a Cycle-GAN,  we also use a ``standard'' Cycle-GAN (StaCyc)\footref{foot1}. To compare against other methods, we modified code from \cite{wang20con} comparing the same methods excluding CIDA, since it behaved worse or comparable to the PCIDA variant in \cite{wang20con}. 
To train the Cycle-GAN, we followed standard practices\cite{zhu17}, i.e., we assessed $\lambda=\{1,3,10\}$ and found somewhat better results for $\lambda=3$. We used the Adam Optimizer decaying the learning rate for the last 1/3 of epochs. We conducted 12 runs and reported not the median (50\% quantile) but the quartile (25\% quantile), since we found that the Cycle-GAN sometimes does not converge properly. Both the median and quartile are more robust against outliers.  




\begin{table}
    \setlength\tabcolsep{3.5pt}
    \caption{Results Fashion: Overall, DoTra outperforms on rotation and is close to the best or second best method otherwise. Note, DoTra predicts samples for target domains $T_1$ and $T_2$, while other methods require samples to be given.}\label{tab:results2} 
	\centering
	\scriptsize
	\begin{tabular}{|r| r|r|r| r| c|	} \hline 
Method/& Source           & Target& Target & Target & \footnotesize{\textbf{Given Target}}   \\
\emph{Operation}&  Domain &Dom. 0& Dom. 1&Dom. 2 & \footnotesize{\textbf{Domains}}\\ \hline 
\multicolumn{1}{|l|}{\emph{Zoom}}&\multicolumn{1}{|l|}{\emph{ 0 }}&\multicolumn{1}{|l|}{\emph{ 1.33 }}&\multicolumn{1}{|l|}{\emph{ 1.77 }}&\multicolumn{1}{|l|}{\emph{ 2.35}} & \\ \hline
 \textbf{DoTra}\tiny{(Our)} &92.2&69.3&61.1&48.5&\multirow{2}{*}{\small{\textbf{only 0}}}\\ \cline{1-5}
StaCyc& 92.2&81.4&62.9&42.1&\\ \hline
Source & 92.4&71.5&48.1&30.4& -\\ \hline
ADDA&91.6&70.9&51.4&42.4&\multirow{5}{*}{\small{\textbf{0, 1 and 2}}}\\ \cline{1-5}
CUA&84.2&70.6&59.6&48.7&\\ \cline{1-5}
DANN&90.1&71.8&54.8&49.0&\\ \cline{1-5}
PCIDA&91.2&61.8&49.8&41.8&\\ \cline{1-5}
SO&92.0&57.7&25.5&18.6&\\ \hline \hline

\hline \multicolumn{1}{|l|}{\emph{SplitMid}} &\multicolumn{1}{|l|}{\emph{ 0 }}&\multicolumn{1}{|l|}{\emph{ 6 }}&\multicolumn{1}{|l|}{\emph{ 12 }}&\multicolumn{1}{|l|}{\emph{ 18}} & \\ \hline
   \textbf{DoTra}\tiny{(Our)} &92.5&77.7&62.2&38.3&\multirow{2}{*}{\small{\textbf{only 0}}}\\ \cline{1-5}
   StaCyc& 92.4&87.0&78.1&58.0&\\ \hline
   Source & 92.4&68.7&46.5&25.0&-\\ \hline
   
ADDA&91.8&75.4&63.5&50.2&\multirow{5}{*}{\small{\textbf{0, 1 and 2}}}\\ \cline{1-5}
CUA&84.1&73.1&70.1&70.1&\\ \cline{1-5}
DANN&91.4&81.3&75.6&71.1&\\ \cline{1-5}
PCIDA&91.2&68.0&53.2&40.4&\\ \cline{1-5}
SO&92.2&50.2&31.8&21.3&\\ \hline \hline

\hline \multicolumn{1}{|l|}{\emph{Rotation}} &\multicolumn{1}{|l|}{\emph{ 0 }}&\multicolumn{1}{|l|}{\emph{ 45 }}&\multicolumn{1}{|l|}{\emph{ 90 }}&\multicolumn{1}{|l|}{\emph{ 135 }} &\\ \hline
 \textbf{DoTra}\tiny{(Our)} &92.2&68.3&64.2&62.1&\multirow{2}{*}{\small{\textbf{only 0}}}\\ \cline{1-5}
 StaCyc& 92.4&15.7&14.1&10.0&\\ \hline
Source & 92.4&21.6&8.7&15.6&-\\ \hline
ADDA&89.6&17.9&5.3&33.2&\multirow{5}{*}{\small{\textbf{0, 1 and 2}}}\\ \cline{1-5}
CUA&83.2&28.1&13.3&18.6&\\ \cline{1-5}
DANN&90.3&39.2&30.0&24.8&\\ \cline{1-5}
PCIDA&91.4&62.2&64.8&82.2&\\ \cline{1-5}
SO&92.4&15.2&7.4&13.4&\\\hline

\end{tabular}
	
\end{table}

\noindent\textbf{Results:} Tables \ref{tab:results} and \ref{tab:results2} confirms that DoTra is on-par with the best method for the source domain. Several methods are considerably worse than the baseline denoted as ``Source'' that simply trains on the given labeled data $D_S$ and predicts samples from target domains. DoTra outperforms or is comparable on the first target domain $T_0$ corresponding to a solution to the unsupervised domain adaption. For the second target domain, DoTra predicts samples, while all other methods (except source and the standard Cycle-GAN (StaCyc)) can use target samples. Surprisingly, DoTra manages to outperform these methods on all operations with the exception of rotation for MNIST. For Fashion it outperforms for rotation, is on par for zoom, and is among the top 3 for SplitMid. This is unexpected since these methods can leverage significantly more information, i.e., they can use data from all three target domains. For rotations, PCIDA tends to yield very good results for rotations but does not strictly outperform on all domains. That is, accuracy fluctuates across domains for PCIDA, i.e. it is worse for $T_2$ than $T_1$. This was also reported by the authors of PCIDA. It should be noted that the PCIDA model is not designed to be general. It is tailored to rotations, i.e., the methods perform (as expected) poorly on all other benchmarks. Generally, on the third target domain, $T_2$ results are more mixed. While DoTra consistently performs well, depending on the operation, other techniques leveraging samples from all domains outperform. Overall, we find it surprising that DoTra performs so well even when predicting samples from target domains like $T_1$ and $T_2$ without the benefit of leveraging samples from these domains that other methods enjoy.

\begin{figure*}[!h]
 \centering{  \includegraphics[width=0.7\linewidth]{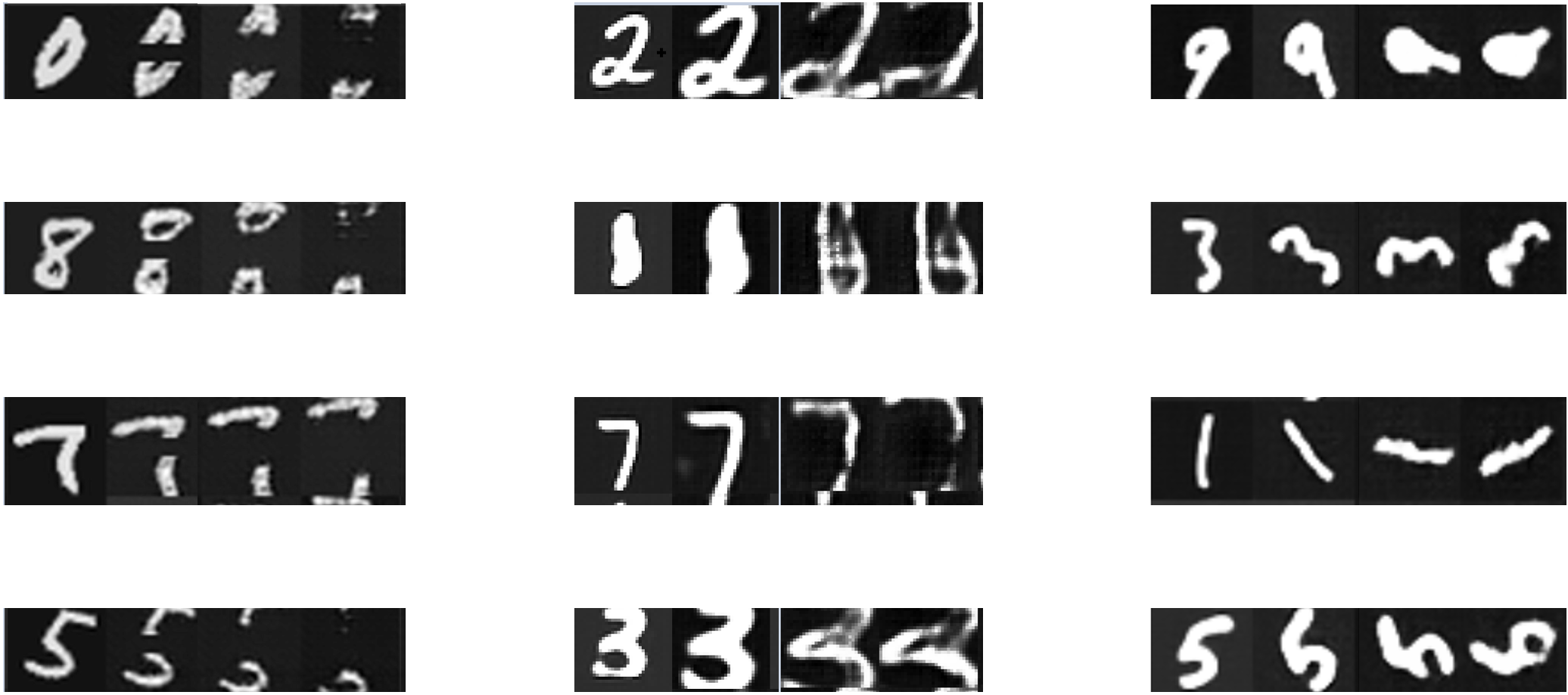}}
 \caption{Samples obtained from the learned transformation $G$ for MNIST. As expected, iterated application of $G$ leads to a distortion of samples.} \label{fig:samples1}
\end{figure*}

\begin{figure}[!h]
 \centering{  \includegraphics[width=0.9\linewidth]{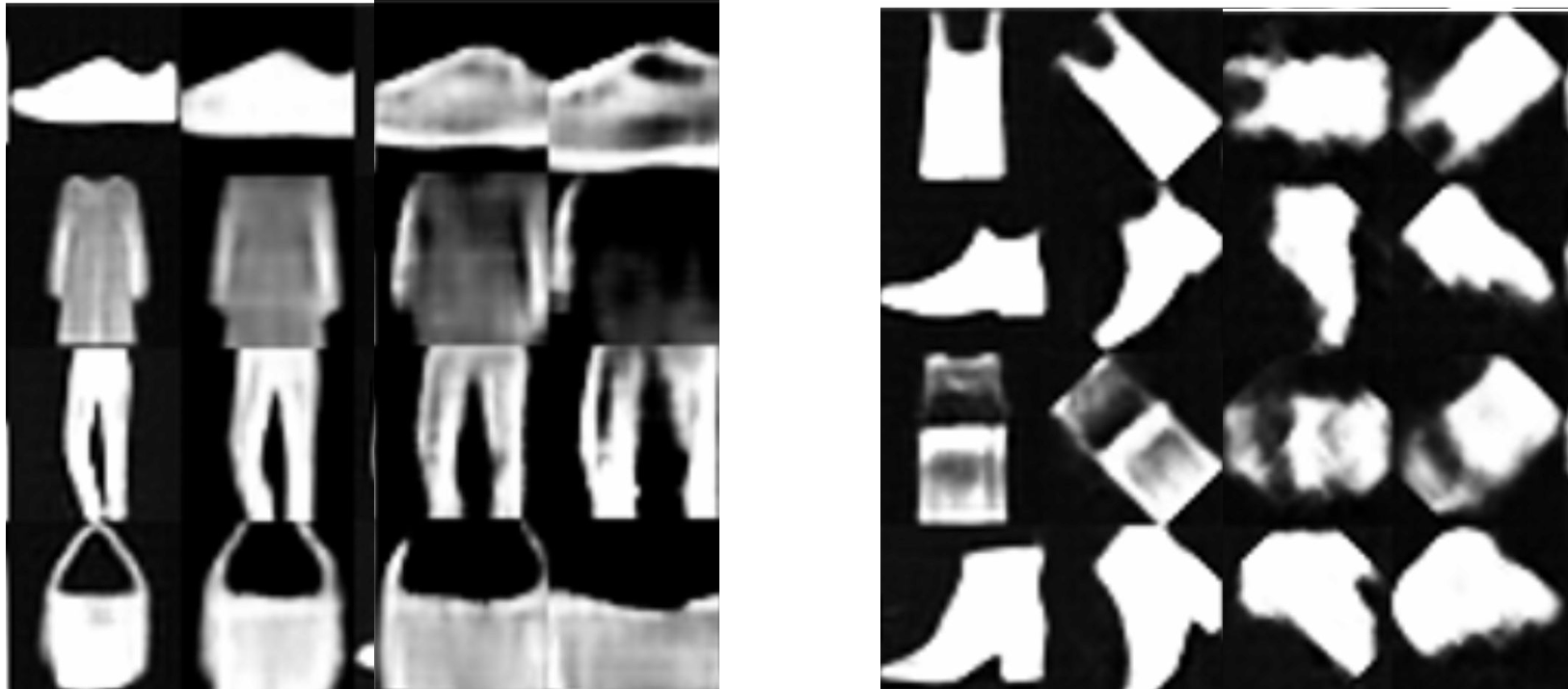}}
 \caption{Samples obtained from the learned transformation $G$ for zoom(left) and rotations(right) for Fashion. As expected, iterated application of $G$ leads to a larger distortion of samples.} \label{fig:samples3}
\end{figure}

However, for DoTra continued application of the transformation worsens results in a non-linear manner. That is, the performance gap between domains $D_{T_i}$ and $D_{T_{i+1}}$ reduces faster than linear in $i$. This also becomes apparent in Figure \ref{fig:samples1} for MNIST and Figure \ref{fig:samples3} for Fashion, where transformations to target domain $T_0$ seem of very high quality, while those of the target domain $T_2$ show more distortion. This also holds for other transformations such as rotation (right panel in Figure \ref{fig:samples1}), where it is not trivial to distinguish some classes, e.g., 4 and 9, 5 and 6 for the target domain $T_2$. For splitting DoTra mainly struggles to fill in the ``blanks'' or close them, i.e., the ``blanks'' are the areas that are newly added due to the splitting, i.e., shifting the upper half upwards and the lower half downwards. For zooming, there is a tendency for the digits to fade rather than to get thicker, while the overall shape is correctly scaled. \\
For rotation, ADDA and DANN often even yield lower accuracy compared to not performing adaptation at all. This has also been reported for rotation in a similar setting in \cite{wang20con} arguing that without capturing the underlying structure, adversarial encoding alignment may harm the transferability of data. The standard Cycle-GAN (StaCyc) performs very well (especially for Fashion) except for rotations, for which it fails for both datasets. Next, we investigate this further using additional analysis.

\begin{figure}[!h]
 \centering{  \includegraphics[width=0.6\linewidth]{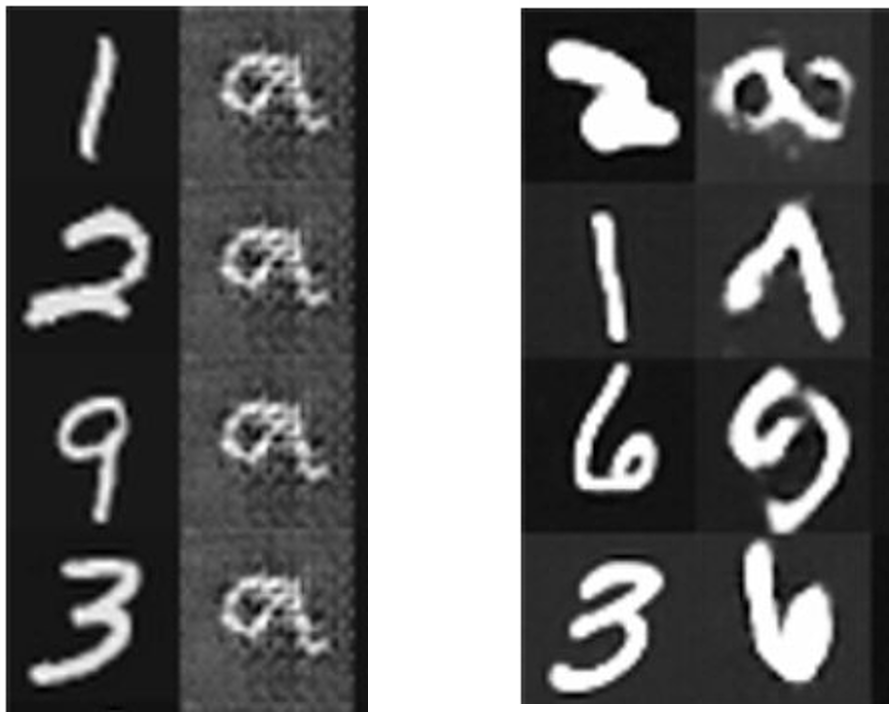}}
 \caption{Cycle-GANs might fail to properly match target domains as illustrated for rotations. The left panel shows mode collapse with the original samples in the left column and transformed in the right. The right panel shows ``sample''(class) switching, where original samples (left column) are mapped to samples of different classes (right).} \label{fig:err}
\end{figure}

\subsection{Further Analysis of DoTra vs StaCyc}
The results for DoTra vs StaCyc are intriguing since there is a huge performance gap for rotations but not for other operations, although except for the applied transformation $G^u$ to the data, the setups are identical among operations.\\
First of all, we found that training can be instable for both. Failures occur due to mode collapse and switching samples from different classes -- illustrated in Figure \ref{fig:err}. We found that mode collapse is not occurring or rare for DoTra but more common when training directly to transform between domains. However, both DoTra and the ordinary Cycle-GAN commonly suffer from switching samples. Reflecting on the differences between DoTra and StaCyc led to the following propositions: \\
\emph{Proposition 1:} Source and target domain appear to be more overlapping for rotation than for other operations. For example, a rotated 1 might appear as a 7 (and vice versa). This makes it more difficult for the discriminator to identify fake images and also makes learning a simple identity transformation by the generator more attractive.\\
\emph{Proposition 2:} Rotation is difficult to express in terms of convolutions. That is, there are only a few solutions for the parameters of the network that yield good performance. 

Shifting (= translating) is very easy to express in a CNN. For example,  to shift by one pixel, a point at location $(i,j)$ is moved to $(i+1,j)$. A single 3x3 filter suffices, i.e.
$\tiny \begin{pmatrix} \label{mat}
0&0&0\\
0&0&1\\
0&0&0\\
\end{pmatrix}$. 
\emph{Zoom}ing is more involved in terms of the number of filters needed and location dependency. That, is a point at location $(i,j)$ is moved to $(s\cdot i,s\cdot j)$ using a simple scaling matrix. That is, every location is ``shifted'' by a different amount depending on location.
For rotation, also every location is moved differently, i.e., to rotate by an angle $\theta$, we move coordinates $(i,j)$ to $(i\cos\theta-j\sin\theta, i\sin\theta+j\cos\theta)$ which follows from using a standard rotation matrix. 
Comparing how much coordinates are moved for zooming with $s=1.33$ and rotation for $\theta=45$ degrees,  yields for zooming $(i,j)-(s\cdot i,s\cdot j)=(0.33i,0.33j)=i\cdot (0.33,0.33)$. For rotation we get using  $\sin 45=\cos 45 = 0.7$: $(i,j)-(i\cos\theta-j\sin\theta,i\sin\theta+j\cos\theta)=(0.3i-0.7j,0.7i-0.3j)$. To simplify, let us set $j=0$ to get $(0.3i,0.7i)$. Thus, rotation moves at least some of the pixels substantially further than zooming.\footnote{Note, some points of the original image are cropped after rotation, e.g., all corners.} Moving points for larger distances is more challenging than moving for shorter distances. The reason is the CNN architecture, where the receptive field of a filter in a layer of the CNN essentially limits the options the network has at its avail to learn to move a point.

To assess Propositions 1 and 2 empirically, we performed the following tests: (a) We trained with rotating only by half the angle, e.g. 22.5 degrees. This reduces the distance points are moved. (b) We employed additional layers for the transformer, i.e., for the architecture in Table \ref{tab:arch} we considered a network with generator $G^{+1}$, where we added one extra 3x3 conv layer with stride 1 after the first downsampling layer and one extra 3x3 deconv layer with stride 1 before the last upsampling layer. This increases considerably the possible number of network parameter settings that yield good transformers.
The results for the standard Cycle-GAN (StaCyc) for rotation are shown in Table \ref{tab:resSta}. Results for using $G^{+}$ and rotations by 22.5 degrees are comparable for both datasets. For Fashion the results considerably improve, when rotating less or increasing the receptive field. Thus, the findings support Proposition 2. For MNIST, results did not strongly change when changing the rotation or architecture. This is supportive of Proposition 1, saying that the main problem is not that the set of (good) solutions for parameters is small, but the overlap of domains.

\begin{table}
    \setlength\tabcolsep{3.5pt}
	\centering
	\scriptsize
	\caption{Results for Proposition Testing on Standard Cycle-GAN.}\label{tab:resSta} 
	\begin{tabular}{|l|l| r|r|r| r| 	} \hline 
Data&Setup& Source           & Target& Target & Target    \\
&&  Domain &Dom. 0& Dom. 1&Dom. 2 \\ \hline 

Fashion&Standard Cycle-GAN (StaCyc) & 92.4&15.7&14.1&10.0\\ \hline
Fashion&StaCyc with $G^{+1}$ & 92.3&82.2&67.0&61.1\\ \hline
Fashion&StaCyc with 22.5 degree rot.& 91.9&84.4&77.1&68.9\\ \hline \hline

MNIST&StaCyc & 99.5&47.4&21.9&13.6\\ \hline
MNIST& StaCyc with $G^{+1}$ & 99.5&44.9&24.8&17.2\\ \hline
MNIST&StaCyc with 22.5 degree rot. & 99.6 & 43.1&41.9&23.3\\ \hline
\end{tabular}
	
 \end{table}

\section{Related Work}
\paragraph{Unsupervised domain adaption} Domain adaption is well studied - see \cite{wil20} for a recent survey with applications in many domains such as health care \cite{chou20} and autonomous driving \cite{trie21}. The machine learning problems comprise of image recognition, e.g., \cite{wang20con}, but also other problems such as image segmentation, e.g., \cite{lee2021}, object detection, e.g., \cite{li2021cat}, and activity recognition \cite{san21}. Domain adaption might also leverage multiple data sources, e.g., \cite{wil20a}. However, most of these works do not aim at dealing with or extrapolating to domains for which no samples are available. 
\paragraph{Domain extrapolation} Some works on domain adaption claim to ``generalize to unseen domains'' \cite{wang21}. From our perspective, this is a misnomer, since only labels are unseen, but samples are actually seen. However, a few works also dealt with domain extrapolation, where a classifier was assessed on samples of a domain it has not seen any samples from. Most notably,  \cite{wang20con} investigated continuous domain adaption, where domains are seen as evolving over time. They relied on prior techniques, i.e., adversarial adaptation, to generate invariant features for a single classifier, altering the discriminator and regressing on the index. To learn domains originating through rotation they explicitly modeled the rotation operation using a spatial transformer network. However, anticipating the transformation and deriving explicit models for it, is labor-intensive, difficult, and against the spirit of machine learning. In contrast, we use a generic architecture that is not limited to rotations only. \cite{bob18} performs continuous adaption under the assumption that domain adaption is continuous and gradual. They continuously update a classifier from one (discretized) domain to the next. \cite{kumar20} focuses on gradual domain adaptation. That is, samples (from ``different'' domains) must bear large similarities so that the proposed pseudo labeling succeeds. In contrast, we allow for larger differences between domains. 
\paragraph{Cycle-GANs for domain adaption}
\cite{ric20} used linear, orthogonal transformations to map images from one domain to another domain. They argued that many existing approaches, e.g., based on Cycle-GANs, are restricted to learning local transformation, but fail to learn non-local ones, such as flipping of images. \cite{alm18} suggests an augmented Cycle-GAN, which introduces a latent variable in the source and target domain in addition to the transformed samples. This allows to learn a many to many mapping, e.g. one source image might result in a multitude of target images. In our case, we aim to learn a 1:1 mapping, since a source image should be transformed using a deterministic function. However, in future work, one might also consider transformations using a stochastic function, where the approach of \cite{alm18} could be of great value.
\paragraph{Autoencoder and domain adaption}
Autoencoders are also commonly used for domain adaption. The idea of using encodings in an adversarial setting has been used before for domain adaptation but also for GANs and AEs in general \cite{li2015gen,sch22con}. \cite{pid20} uses two AEs, i.e., an MLP and StyleGAN. They learn disentangled representations leading to similar quality images as GAN only based images. \cite{yan19} performed multi-domain translation using AE. That is, they assumed a shared latent representation across all domains. In contrast, our AE yields typically non-shared domain latent encodings. We learn a transformer to translate between two latent spaces. Ideas from \cite{yan19} might be valuable, if we are given samples from more domains than just the source and one target domain. \cite{par20} focused on image manipulation. They used AEs to learn concurrently a structural and texture code. Codes of different images can be mixed during reconstruction. \cite{hou18} assumes that images of different domains have the same latent code for content but not for style. \cite{zho19} trained a Cycle-GAN based on a shared encoder on the source and target domain, but separate decoders and multiple loss terms, including, e.g., a classification loss. In contrast, we use the Cycle-GAN only on encoded, low-dimensional data. \cite{zhao2017} considers a specific problem for sleep stages. Their aim is to remove conditional dependencies rather than making the representation domain independent. Their adversary used the representation but it is also conditioned on the predicted label distribution.





\section{Discussion, Limitations and Future Work}
In this work, we advocate the learning of a transformation to predict unseen samples. Our solution allows dealing with problems such as concept drift in a proactive manner rather than reacting once they have occurred. \\
We relied on Cycle-GANs that are non-easy to train and might not converge to an optimal solution. That is, we found that Cycle-GANs might mismatch classes between domains, e.g. the class ``pants'' in the source domain might become class ``shoes'' in the target domain. Other approaches, e.g. based on graphs and clustering samples based on latent representations, and computing a matching between graphs of the source and target domain will be investigated in future work. One might also leverage other approaches from unsupervised domain adaption than Cycle-GANs such as semi-supervised learning and learning shared representations.  They have been shown to work well when domains are similar (from the perspective of the investigated network). For example, state-of-the-art CNN architectures are generally not rotation invariant\cite{zei14}. Thus, images with large rotations appear very different from these networks. They can imply that accuracy for multiple classes can drop from being mostly correct, e.g., above 80\% accuracy, to being mostly incorrect, e.g., accuracy below 10\%\cite{zei14}. In these situations, our approach might yield the largest benefits compared to other approaches. Though, we conjecture that a hybrid approach is probably the best.\\
It might also be interesting to apply XAI techniques to better understand the learnt transformations, e.g., by visualizing different layers of the transformer \cite{sch21,sch22exp}. Further improving the architecture of DoTra, e.g., by including novel insights on representation learning\cite{sch21loc}, is also an interesting future direction.


Our work also suggests possible design implications for Cycle-GANs, showing that the receptive field should be carefully assessed. Depending on the training and the data it can be a problem for standard Cycle-GANs but not for DoTra. Training a Cycle-GAN on latent dimensions is faster, more stable, and less prone to mode collapse.


\section{Conclusions}
This work proposes a proactive approach to deal with shifts in data distribution, i.e., domains evolving over time. We obtain a domain transformer that can be applied iteratively to get labeled samples in any (future) target domain where samples are not yet available. Our approach is also beneficial for the classical problem of unsupervised domain adaption, where target samples of a domain (but not labels) are available.
\bibliographystyle{IEEEtran}
\bibliography{refs}

\begin{thebibliography}{10}
\providecommand{\url}[1]{#1}
\csname url@samestyle\endcsname
\providecommand{\newblock}{\relax}
\providecommand{\bibinfo}[2]{#2}
\providecommand{\BIBentrySTDinterwordspacing}{\spaceskip=0pt\relax}
\providecommand{\BIBentryALTinterwordstretchfactor}{4}
\providecommand{\BIBentryALTinterwordspacing}{\spaceskip=\fontdimen2\font plus
\BIBentryALTinterwordstretchfactor\fontdimen3\font minus
  \fontdimen4\font\relax}
\providecommand{\BIBforeignlanguage}[2]{{%
\expandafter\ifx\csname l@#1\endcsname\relax
\typeout{** WARNING: IEEEtran.bst: No hyphenation pattern has been}%
\typeout{** loaded for the language `#1'. Using the pattern for}%
\typeout{** the default language instead.}%
\else
\language=\csname l@#1\endcsname
\fi
#2}}
\providecommand{\BIBdecl}{\relax}
\BIBdecl

\bibitem{lu18}
J.~Lu, A.~Liu, F.~Dong, F.~Gu, J.~Gama, and G.~Zhang, ``Learning under concept
  drift: A review,'' \emph{IEEE Transactions on Knowledge and Data
  Engineering}, 2018.

\bibitem{zhu17}
J.-Y. Zhu, T.~Park, P.~Isola, and A.~A. Efros, ``Unpaired image-to-image
  translation using cycle-consistent adversarial networks,'' in
  \emph{Proceedings of the IEEE international conference on computer vision},
  2017, pp. 2223--2232.

\bibitem{wang20con}
H.~Wang, H.~He, and D.~Katabi, ``Continuously indexed domain adaptation,'' in
  \emph{International Conference on Machine Learning}.\hskip 1em plus 0.5em
  minus 0.4em\relax PMLR, 2020, pp. 9898--9907.

\bibitem{wil20}
G.~Wilson and D.~J. Cook, ``A survey of unsupervised deep domain adaptation,''
  \emph{ACM Transactions on Intelligent Systems and Technology (TIST)},
  vol.~11, no.~5, pp. 1--46, 2020.

\bibitem{chou20}
A.~Choudhary, L.~Tong, Y.~Zhu, and M.~D. Wang, ``Advancing medical imaging
  informatics by deep learning-based domain adaptation,'' \emph{Yearbook of
  medical informatics}, vol.~29, no.~01, pp. 129--138, 2020.

\bibitem{trie21}
L.~T. Triess, M.~Dreissig, C.~B. Rist, and J.~M. Z{\"o}llner, ``A survey on
  deep domain adaptation for lidar perception,'' in \emph{2021 IEEE Intelligent
  Vehicles Symposium Workshops (IV Workshops)}, 2021, pp. 350--357.

\bibitem{lee2021}
S.~Lee, J.~Hyun, H.~Seong, and E.~Kim, ``Unsupervised domain adaptation for
  semantic segmentation by content transfer,'' in \emph{AAAI Conf. on
  Artificial Intelligence}, 2021.

\bibitem{li2021cat}
S.~Li, J.~Huang, X.-S. Hua, and L.~Zhang, ``Category dictionary guided
  unsupervised domain adaptation for object detection,'' in \emph{Proceedings
  of the AAAI Conference on Artificial Intelligence}, vol.~35, no.~3, 2021, pp.
  1949--1957.

\bibitem{san21}
A.~R. Sanabria, F.~Zambonelli, and J.~Ye, ``Unsupervised domain adaptation in
  activity recognition: A gan-based approach,'' \emph{IEEE Access}, vol.~9, pp.
  19\,421--19\,438, 2021.

\bibitem{wil20a}
G.~Wilson, J.~R. Doppa, and D.~J. Cook, ``Multi-source deep domain adaptation
  with weak supervision for time-series sensor data,'' in \emph{Proceedings of
  the 26th ACM SIGKDD International Conference on Knowledge Discovery \& Data
  Mining}, 2020, pp. 1768--1778.

\bibitem{wang21}
J.~Wang, C.~Lan, C.~Liu, Y.~Ouyang, W.~Zeng, and T.~Qin, ``Generalizing to
  unseen domains: A survey on domain generalization,'' \emph{arXiv:2103.03097},
  2021.

\bibitem{bob18}
A.~Bobu, E.~Tzeng, J.~Hoffman, and T.~Darrell, ``Adapting to continuously
  shifting domains,'' in \emph{Workshop at International Conference on Learning
  Representations}, 2018.

\bibitem{kumar20}
A.~Kumar, T.~Ma, and P.~Liang, ``Understanding self-training for gradual domain
  adaptation,'' in \emph{International Conference on Machine Learning}, 2020.

\bibitem{ric20}
E.~Richardson and Y.~Weiss, ``The surprising effectiveness of linear
  unsupervised image-to-image translation,'' \emph{arXiv preprint
  arXiv:2007.12568}, 2020.

\bibitem{alm18}
A.~Almahairi, S.~Rajeshwar, A.~Sordoni, P.~Bachman, and A.~Courville,
  ``Augmented cyclegan: Learning many-to-many mappings from unpaired data,'' in
  \emph{International Conference on Machine Learning}, 2018, pp. 195--204.

\bibitem{li2015gen}
Y.~Li, K.~Swersky, and R.~Zemel, ``Generative moment matching networks,'' in
  \emph{International Conference on Machine Learning}.\hskip 1em plus 0.5em
  minus 0.4em\relax PMLR, 2015, pp. 1718--1727.

\bibitem{sch22con}
J.~Schneider and G.~Apruzzese, ``Concept-based adversarial attacks: Tricking
  humans and classifiers alike,'' \emph{arXiv preprint arXiv:2203.10166}, 2022.

\bibitem{pid20}
S.~Pidhorskyi, D.~A. Adjeroh, and G.~Doretto, ``Adversarial latent
  autoencoders,'' in \emph{Proc. of Conference on Computer Vision and Pattern
  Recognition}, 2020.

\bibitem{yan19}
K.~Yang and C.~Uhler, ``Multi-domain translation by learning uncoupled
  autoencoders,'' in \emph{Computational Biology Workshop, Int. Conference on
  Machine Learning}, 2019.

\bibitem{par20}
T.~Park, J.-Y. Zhu, O.~Wang, J.~Lu, E.~Shechtman, A.~A. Efros, and R.~Zhang,
  ``Swapping autoencoder for deep image manipulation,''
  \emph{arXiv:2007.00653}, 2020.

\bibitem{hou18}
H.~Hou, J.~Huo, and Y.~Gao, ``Cross-domain adversarial auto-encoder,''
  \emph{arXiv preprint arXiv:1804.06078}, 2018.

\bibitem{zho19}
Q.~Zhou, B.~Yang \emph{et~al.}, ``Deep cycle autoencoder for unsupervised
  domain adaptation with generative adversarial networks,'' \emph{IET Computer
  Vision}, 2019.

\bibitem{zhao2017}
M.~Zhao, S.~Yue, D.~Katabi, T.~S. Jaakkola, and M.~T. Bianchi, ``Learning sleep
  stages from radio signals: A conditional adversarial architecture,'' in
  \emph{International Conference on Machine Learning}, 2017, pp. 4100--4109.

\bibitem{zei14}
M.~D. Zeiler and R.~Fergus, ``Visualizing and understanding convolutional
  networks,'' in \emph{European conference on computer vision}, 2014, pp.
  818--833.

\bibitem{sch21}
J.~Schneider and M.~Vlachos, ``Explaining neural networks by decoding layer
  activations,'' in \emph{International Symposium on Intelligent Data
  Analysis}, 2021, pp. 63--75.

\bibitem{sch22exp}
------, ``Explaining classifiers by constructing familiar concepts,''
  \emph{Machine Learning}, pp. 1--34, 2022.

\bibitem{sch21loc}
J.~Schneider, ``Locality-promoting representation learning,'' in
  \emph{International Conference on Pattern Recognition (ICPR)}, 2021, pp.
  8061--8068.

\end{thebibliography}

\end{document}